\documentclass[runningheads]{llncs}

\usepackage{amsmath,amssymb,amsfonts}
\usepackage{algorithm}
\usepackage{algorithmic}
\usepackage{graphicx}
\usepackage{textcomp}
\usepackage{xcolor}
\usepackage{booktabs}
\usepackage{array}
\usepackage{multirow}
\usepackage{tabularx}
\usepackage{url}
\usepackage{xspace}

\newcommand{\method}{HaReCAP\xspace}

\newcommand{\figplaceholder}[2]{%
\IfFileExists{#2.pdf}{%
  \includegraphics[width=\linewidth]{#2.pdf}%
}{%
  \IfFileExists{#2.png}{%
    \includegraphics[width=\linewidth]{#2.png}%
  }{%
    \IfFileExists{#2.eps}{%
      \includegraphics[width=\linewidth]{#2.eps}%
    }{%
      \fbox{\begin{minipage}{0.92\linewidth}
      \centering
      \vspace{0.6em}
      \textbf{#1}\\[0.3em]
      {\footnotesize Missing figure: \texttt{#2}}
      \vspace{0.6em}
      \end{minipage}}%
    }%
  }%
}%
}

\begin{document}

\title{\method: Habitual-action Grounding for Recursive Large Language Model Agents}

\titlerunning{\method: Habitual-action Grounding}

\author{
Shen Liu\inst{1} \and
Zhenguo Xu\inst{2} \and
Shaopu Wang\inst{3} \and
Yike Gao\inst{3} \and
Chunlei Wang\inst{3}\thanks{Corresponding author.}
}

\authorrunning{S. Liu et al.}

\institute{
North China Institute of Computer System Engineering, Beijing, China\\
\email{liushen24@mails.ucas.edu.cn}
\and
University of Science and Technology of China, Hefei, China\\
\email{xzg@mail.ustc.edu.cn}
\and
China Information Security Research Institute Co., Ltd., Beijing, China\\
\email{wangshaopu18@mails.ucas.ac.cn, yikegao@mail.ustc.edu.cn, wangchl@cecr.ac.cn}
}



\maketitle

\begin{abstract}
Long-horizon embodied tasks require LLM agents to iteratively decompose high-level goals, revise plans in response to environmental feedback, and ground leaf-level subgoals into valid executable actions.
Recursive context-management methods such as ReCAP improve planning stability through multi-level task decomposition and parent-node refinement, but still repeatedly invoke the LLM at leaf nodes to ground atomic subtasks into exact valid actions.
We refer to this final grounding step as \emph{last-mile grounding redundancy}, which accumulates into substantial LLM-call and token overhead during long-horizon execution.
To mitigate this issue, we propose \method \footnote{Code and data: https://github.com/agic01/harecap} (Habitual-action Grounded ReCAP), a low-intrusion leaf grounding extension for ReCAP.
\method extracts frequent leaf decisions from successful trajectories and compiles them offline into auditable and abstainable one-step leaf-reflex rules. 
At runtime, it skips the leaf LLM call only when a rule can uniquely determine a legal action in the current valid-action set; otherwise, it falls back to the original ReCAP.
This design avoids repeatedly carrying the full recursive context into the LLM for routine leaf action grounding, while preserving the original recursive control flow.
We evaluate \method  on Robotouille and ALFWorld with Qwen3.5-27B as the main model. 
On tasks solved by both ReCAP and \method, \method reduces token consumption by 14.67\%, 17.93\%, and 20.08\% on Robotouille synchronous, Robotouille asynchronous, and ALFWorld, respectively.
The results show that \method can serve as a low-intrusion extension to ReCAP-style recursive context-management frameworks, reducing \emph{last-mile grounding redundancy} across environments and models on commonly successful trajectories.

\end{abstract}

\section{Introduction}
Long-horizon embodied tasks require large language model (LLM) agents to perceive the environment, decompose goals, update plans, and produce executable actions over continuous interaction.
Unlike one-shot question answering or short-form reasoning, these tasks expose the model to changing observations, local subgoals, and valid-action constraints.Consequently, the agent must maintain the consistency of the goal and the legality of the action in a dynamic context.
Chain-of-Thought (CoT) and Tree-of-Thoughts (ToT) highlight the importance of explicit reasoning processes and search-based reasoning strategies for improving complex task-solving capabilities \cite{cot,tot}. Extending this line of work, ReAct couples reasoning with action execution, allowing models to iteratively update their decisions through environmental feedback, thereby improving performance in interactive decision-making tasks \cite{react}.
Recently, ReCAP~\cite{recap}, a state-of-the-art framework for long-horizon execution, introduces recursive context management to support complex task completion.
It organizes execution through plan-ahead decomposition, parent-plan re-injection, and memory-efficient execution: the model generates a subtask list at the current node, the system recursively descends into the first subtask, and after that subtask is completed, the remaining parent plan is re-injected and refined with the latest observation.
This mechanism avoids naive linear accumulation of the full interaction history and has shown clear advantages on long-horizon benchmarks such as Robotouille \cite{robotouille}.

However, ReCAP\cite{recap} does not remove all repeated reasoning.
At each leaf node, the model still has to translate a natural-language subtask into an action string that exactly matches the current valid-action set.
For example, once recursive decomposition has produced the leaf subtask ``pick up onion1 from board1'', the leaf LLM call often only grounds it into the exact valid action ``Pick up onion1 from board1 using robot1''.
Although this step appears simple, it still carries the ReCAP\cite{recap} dialogue history, parent-task context, latest observation, and output JSON schema, consuming another round of prompt and completion tokens. 
This leaf-level action-grounding overhead is repeatedly amplified, since long-horizon tasks contain many atomic actions.
We call this repeated reasoning process at the end of recursive planning \emph{last-mile grounding redundancy}.
As shown in Fig.~\ref{fig:last-mile-cost}, among successful Qwen3.5-27B ReCAP\cite{recap} trajectories, strict leaf/action grounding calls account for 23.67\%, 29.56\%, and 45.73\% of all LLM calls on Robotouille\cite{robotouille} synchronous, Robotouille\cite{robotouille} asynchronous, and ALFWorld\cite{alfworld}, respectively.
This indicates that last-mile grounding is not an incidental edge cost, but a stable component of inference cost in recursive LLM agents.

\begin{figure}[t]
\centering
\scriptsize
\setlength{\tabcolsep}{3pt}
\begin{tabular}{@{}lcc@{}}
\toprule
Benchmark & ReCAP call composition & Grounding \\
\midrule
Robotouille Sync &
\textcolor{blue!70!black}{\rule{0.300\linewidth}{6pt}}\textcolor{orange!85!black}{\rule{0.093\linewidth}{6pt}} &
23.67\% \\
Robotouille Async &
\textcolor{blue!70!black}{\rule{0.276\linewidth}{6pt}}\textcolor{orange!85!black}{\rule{0.116\linewidth}{6pt}} &
29.56\% \\
ALFWorld &
\textcolor{blue!70!black}{\rule{0.213\linewidth}{6pt}}\textcolor{orange!85!black}{\rule{0.180\linewidth}{6pt}} &
45.73\% \\
\bottomrule
\end{tabular}
\vspace{0.2em}

\scriptsize
\textcolor{blue!70!black}{\rule{1.1em}{5pt}} recursive reasoning/refinement \quad
\textcolor{orange!85!black}{\rule{1.1em}{5pt}} strict leaf/action grounding
\caption{Last-mile action-grounding redundancy in successful ReCAP trajectories. Orange denotes strict leaf/action grounding calls, while blue denotes the remaining recursive reasoning, decomposition, and refinement calls. Statistics are collected from the Qwen3.5-27B ReCAP baseline.}

\label{fig:last-mile-cost}
\end{figure}

\begin{figure}[t]
\centering
\includegraphics[width=0.84\linewidth]{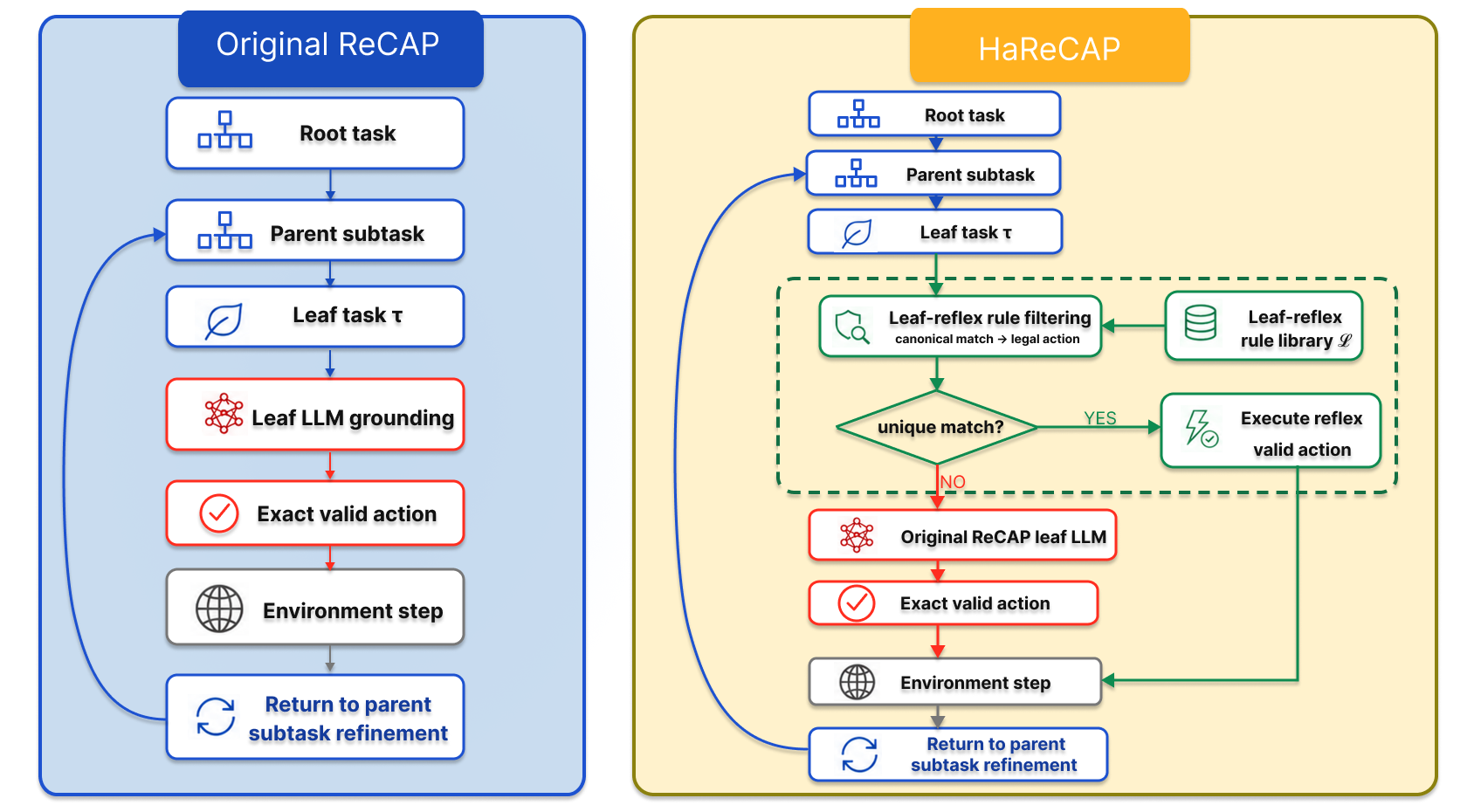}
\caption{Leaf-level difference between original ReCAP and \method. \method preserves recursive planning, backtracking, and refinement, and only inserts a leaf-reflex library at leaf-level action grounding.}
\label{fig:recap-vs-skill}
\end{figure}

A direct alternative is to learn intermediate macro skills, such as compiling ``move onion from fryer to board'' into a sequence of pick-up, move, and place actions.
However, in a recursive framework such as ReCAP\cite{recap}, mid-level macro skills introduce additional risks: they must resolve object bindings, estimate subgoal progress, and decide skill termination.
Direction-sensitive actions such as ``Stack A on B'' versus ``Stack B on A'' can fail when object roles are reversed; under layout changes, a mid-level macro skill may also execute before its preconditions are satisfied.
These issues can alter ReCAP\cite{recap}'s stable recursive decomposition, backtracking refinement, and failure-handling logic.

To avoid these risks, this paper adopts a more focused and more predictable design: distilling only the final action-grounding step at leaf nodes.
We propose \method, Habitual-action Grounded ReCAP.
As summarized in Fig.~\ref{fig:recap-vs-skill}, \method preserves ReCAP\cite{recap}'s recursive task tree and high-level planning, and inserts a local, abstainable, one-step habitual-action grounding module at leaf-level action grounding.
To avoid terminology ambiguity, we use \emph{macro skill} to refer to a multi-step process at an intermediate node, such as a subroutine that continuously executes pick-up, move, and place; 

A \emph{leaf-reflex rule} in \method denotes a one-step mapping from a canonical leaf task to a canonical action template.
All leaf-reflex rules form a \emph{leaf-reflex library}, which is built offline from successful ReCAP\cite{recap} trajectories and is not updated online.
\emph{Habitual-action grounding} refers to the runtime process of retrieving rules from the leaf-reflex library and uniquely instantiating a legal action from the current valid-action set.
The system first extracts leaf tasks, current valid actions, and executed actions from successful ReCAP\cite{recap} baseline trajectories, and compiles them offline into the leaf-reflex library.
During online execution, if the current leaf task can be uniquely grounded by a library rule and the current valid actions, the system directly executes the action; otherwise, it fully falls back to ReCAP\cite{recap}.

The core claim of this design is:
\emph{habitual actions need not replace recursive reasoning; moving frequent leaf-level action grounding out of LLM calls can reduce token cost while preserving the reliability of ReCAP's task decomposition.}
\method is neither a macro-skill planner nor an online episodic memory system.
It does not change ReCAP\cite{recap}'s recursive decomposition, backtracking refinement, or failure-handling logic.
It is closer to last-mile reasoning distillation: ReCAP\cite{recap}'s repeatedly successful final-step action choices are distilled into local reflexes that remain strictly constrained by the current valid-action set.

The contributions of this paper are as follows:
\begin{itemize}
    \item We identify last-mile grounding redundancy in ReCAP-style recursive context-management frameworks, showing that repeated calls from atomic subtasks to exact valid actions constitute an independently optimizable inference cost.
    \item We propose habitual-action grounding, which distills frequent successful leaf decisions into abstainable one-step leaf-reflex rules, retaining experience reuse while avoiding macro-skill takeover risks.
    \item We design a leaf-reflex library construction procedure and a conservative triggering mechanism, so a habitual action is executed only when the current subtask can be uniquely grounded to a legal action, yielding a low-intrusion integration aligned with the original ReCAP\cite{recap} control flow.
    \item We evaluate the mechanism on Robotouille\cite{robotouille} and ALFWorld\cite{alfworld}, separate end-to-end performance from paired-success efficiency, and show that \method reduces LLM calls and token consumption on commonly solved trajectories across environments.
\end{itemize}






\section{Related Work}

\subsection{Reasoning and Planning for LLM Agents}

The core challenge for LLM agents is not merely generating a natural-language plan, but maintaining consistency among observations, local goals, and executable actions.
Chain-of-Thought\cite{cot}, Tree-of-Thoughts\cite{tot}, and Graph-of-Thoughts\cite{graph_of_thoughts} improve complex problem solving through explicit reasoning or search structures, but they mainly target static or weakly interactive problems.
ReAct\cite{react} interleaves reasoning and acting so the model can adjust the next action based on environment feedback; SayCan\cite{saycan}, Inner Monologue\cite{inner_monologue}, and LLM-Planner\cite{llm_planner} further emphasize that language plans must be constrained by affordances, feedback, and executability.
In text-based embodied environments such as Robotouille\cite{robotouille}, each action must also exactly match the current valid-action set.
Thus, even when the model has inferred the correct atomic intent, it still needs to convert that intent into the environment's action template.
This paper focuses precisely on this leaf-level grounding cost.

\subsection{Context Management and Structured Memory}

Long-horizon interaction creates context growth, stale observations, and goal drift.
ReCAP\cite{recap} uses a recursive task tree to maintain parent tasks, active subtasks, and backtracking relations, organizing long histories into locally reasoned contexts.
External memory and graph structures address context limits from another direction: RAG\cite{rag} and GraphRAG\cite{graphrag} organize relevant knowledge as retrievable context, while Plan-on-Graph\cite{plan_on_graph} and Graph-Informed Action Generation\cite{gia} explicitly structure states, actions, or historical experience for similar-experience retrieval, loop detection, or limited lookahead planning.
These methods typically support long-horizon decision making by expanding retrievable information or reorganizing planning states.
\method enters at a different point: it does not expand ReCAP\cite{recap}'s global context structure, nor does it rely on scene graphs, embedding retrieval, or environment simulators.
Instead, it compresses repeated leaf-level action grounding decisions from successful ReCAP\cite{recap} trajectories into local, verifiable rules.
Thus, \method focuses on action-grounding redundancy at the end of recursive context management rather than replacing ReCAP\cite{recap}'s long-horizon memory or planning structure.

\subsection{Experience Reuse and Macro Skill Libraries}

Long-term memory and skill reuse can reduce repeated exploration.
Reflexion\cite{reflexion} summarizes failure experience through verbal feedback; Generative Agents\cite{generative_agents}, MemGPT\cite{memgpt}, and Voyager\cite{voyager} demonstrate the role of long-term memory or open-ended skill libraries in continual interaction; experience retrieval in embodied tasks is also commonly used to provide similar historical trajectories or guide action selection \cite{exrap,gia}.
However, the larger the skill granularity, the more the system must decide preconditions, object bindings, interruption timing, and parent-goal completion.
In environments with stacking order, container state, and asynchronous waiting, mid-level macro skills may bypass ReCAP\cite{recap}'s backtracking refinement.
We therefore use a more conservative leaf-reflex rule: each rule represents only a one-step mapping from a previously successful leaf task to a valid action.
If the current task and valid actions cannot uniquely determine an action, the system abstains and falls back to the original ReCAP\cite{recap}.



\section{Method}

\subsection{Problem Formulation and Overview}
Given a root task $g$, a current observation $o_t$, and a valid-action set $A_t$, ReCAP maintains a recursive task tree $\mathcal{T}$.
Each node $v$ corresponds to a natural-language task $\tau_v$, and the model outputs
\begin{equation}
r_v=(\text{think}_v,\text{subtasks}_v).
\end{equation}
Here $r_v$ denotes the structured response for node $v$; $\text{think}_v$ is a concise rationale about the current task, environment state, and decomposition decision, while $\text{subtasks}_v=[\tau_{v,1},\ldots,\tau_{v,n}]$ is an ordered list of subtasks to be recursively processed.
When $\text{subtasks}_v$ contains multiple subtasks, ReCAP descends into the first one.
When the current subtask has been decomposed to an executable level, the model usually outputs a subtask list containing one candidate action, and the system executes it only if $a\in A_t$.
This exact-valid-action constraint guarantees legal interaction with the environment, but it also creates repeated leaf-level grounding cost.
Many leaf tasks are already close to action templates, e.g., ``Pick up bread1 from table2'', yet the model still needs another call to output ``Pick up bread1 from table2 using robot1''.


We define this cost as last-mile action-grounding redundancy in recursive planning and insert a habitual-action grounding module before ReCAP's leaf LLM call:
\begin{equation}
\pi_{\text{H}}(\tau,A_t;\mathcal{L})\rightarrow a\in A_t \ \text{or}\ \bot,
\end{equation}
where $\tau$ is the current leaf task, $A_t$ is the valid-action set, $\mathcal{L}$ is the leaf-reflex library, and $\bot$ denotes abstention.
If $\pi_{\text{H}}$ returns a unique action, the system directly executes that valid action; otherwise it fully falls back to the original ReCAP leaf LLM call.

\begin{figure}[t]
\centering
\figplaceholder{\method system framework}{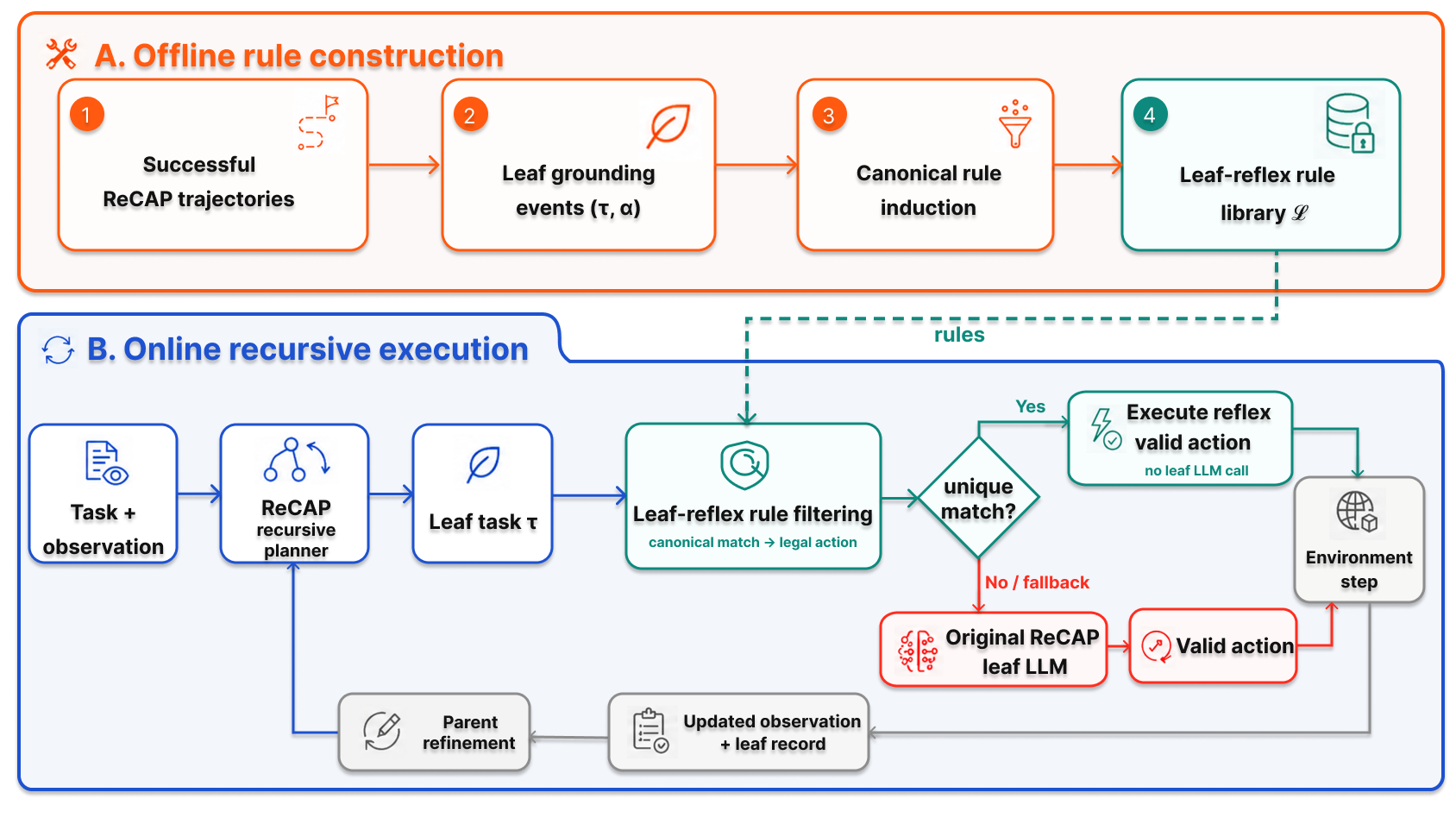}
\caption{System framework of \method. Offline construction builds a leaf-reflex library from successful ReCAP trajectories. Online execution retrieves only before leaf-level action grounding and falls back to the original ReCAP when no rule hits or when ambiguity is detected.}
\label{fig:architecture}
\end{figure}

\subsection{Offline Construction of Leaf-Reflex Rules}



The leaf-reflex library is constructed from successful ReCAP executions. Rather than memorizing complete trajectories or reusable multi-step skills, it retains only the final action-grounding mappings that have been validated through successful execution. When a leaf task is grounded to a valid executable action, the resulting task–action pair is compressed into a leaf-reflex rule:

\begin{equation}
s=(k_\tau,k_a),
\end{equation}
where $k_\tau$ is the canonical key of the leaf-task text and $k_a$ is the canonical key of the executed action.
Intuitively, a rule states that a class of sufficiently concrete leaf tasks should usually be based on a class of action templates; it is not a fixed script for a particular observation.

Rule extraction follows three principles.
First, the extraction granularity is restricted to the local mapping between a leaf task and an executed action; parent subtrees, full task trajectories, and multi-action windows are not compiled into the library.
This prevents the leaf-reflex library from degenerating into a trajectory replay or a macro skill library.
Second, rules are extracted only from successful ReCAP trajectories, ensuring that the library stores one-step grounding decisions already verified by ReCAP rather than unverified intermediate plans.
Finally, support is counted after task-level deduplication: repeated occurrences of the same $(\tau_i,a_i)$ within a single task count only once, preventing loops inside one trajectory from inflating rule confidence.

To improve reusing across object ids and local layouts, we use action-type-agnostic canonicalization rather than hand-written rules for each action class.
This process preserves the word-order structure of tasks and actions, while normalizing object ids into type placeholders, e.g., $\text{bread12}\rightarrow \text{bread\#}$.
Thus leaf tasks and actions are mapped to
\begin{align}
k_\tau &= \texttt{canonical\_leaf\_text}(\tau),\\
k_a &= \texttt{canonical\_leaf\_text}(a).
\end{align}
A candidate event enters the library only if $k_\tau$ and $k_a$ are non-empty, the leaf task explicitly contains object ids, and those ids are covered by the executed action:
\begin{equation}
\mathrm{ids}(\tau)\neq\emptyset,\quad \mathrm{ids}(\tau)\subseteq \mathrm{ids}(a).
\end{equation}
This prevents overly abstract leaf tasks such as ``prepare ingredient'' from being compiled into executable reflexes.

Each leaf-reflex rule is uniquely identified by $(k_\tau,k_a)$ and maintains its occurrence count in successful trajectories as support $n_s$.
In the frozen main experiments, the leaf-reflex library is built offline from successful ReCAP trajectories and is not updated online; runtime triggering is therefore mainly controlled by the support threshold and the unique-action constraint.
The library size and support-threshold ablation are reported in the experimental section.

\subsection{Runtime Retrieval and Triggering}

At runtime, HaReCAP executes the task according to ReCAP's parent-node decomposition, backtracking control, and root-task completion judgment.
The habitual-action grounding module is invoked only immediately before a leaf-level LLM grounding call.
Its inputs are restricted to the current leaf task $\tau$, current valid-action set $A_t$, leaf-reflex library $\mathcal{L}$, and support threshold $m$.
The module does not generate new actions; instead, it checks whether historical rules can uniquely support an already legal action in the current $A_t$.
To avoid triggering on incomplete subgoals, the system first requires $\tau$ to have a non-empty canonical key and explicit object ids.
It then accepts only rules whose task key matches, whose action key can be uniquely instantiated from current valid actions, whose candidate action covers the object ids in the leaf task, and whose support satisfies $n_s\ge m$.
After these filters, each candidate rule $s$ corresponds to at most one valid action in the current environment.
Thus, triggering is primarily governed by strict canonical matching and the unique-valid-action constraint.
When multiple rules pass the hard constraints and map to executable actions, the system uses a support-derived score $\rho(s)$ as a tie-breaker:
\begin{equation}
\rho(s) = 1.6\frac{n_s+1}{n_s+2}+\frac{1}{10}\min(n_s,10).
\end{equation}
The function is monotonic in $n_s$, so higher-support rules are preferred in conflicts.
The first term provides a confidence-like gain that saturates as support increases, while the second term preserves approximately linear separation for $n_s\le 10$.
If two candidates map to different valid actions, the top score must exceed the second score by at least 1; otherwise the candidates are considered within the ambiguity window and the system abstains.
This design allows a low-support rule to be overridden by a clearly higher-support rule, but falls back to ReCAP when two candidates are both reasonably supported and cannot be clearly separated.
This conflict screening is defensive and is not the primary trigger condition.
Algorithm~\ref{alg:leaf-reflex-trigger} formalizes the triggering procedure.
If the algorithm returns $\bot$, HaReCAP takes no local action and falls back to the original ReCAP leaf LLM.

\begin{algorithm}[t]
\caption{Conservative leaf-reflex triggering}
\label{alg:leaf-reflex-trigger}
\small
\begin{algorithmic}[1]
\REQUIRE leaf task $\tau$, valid actions $A_t$, frozen library $\mathcal{L}$, support gate $m$, margin $\lambda=1$
\ENSURE a valid action $a \in A_t$ or abstention $\bot$

\STATE $k_\tau \leftarrow \texttt{canonical\_leaf\_text}(\tau)$
\STATE $I_\tau \leftarrow \mathrm{ids}(\tau)$
\IF{$k_\tau=\emptyset$ or $I_\tau=\emptyset$}
    \RETURN $\bot$
\ENDIF

\STATE $\mathcal{C}\leftarrow \emptyset$
\FORALL{$s \in \mathcal{L}$}
    \IF{$k_\tau^s = k_\tau$ and $n_s \ge m$}
        \STATE $a \leftarrow \texttt{UniqueMatch}(k_a^s, I_\tau, A_t)$
        \IF{$a \neq \bot$}
            \STATE $\mathcal{C}\leftarrow \mathcal{C}\cup\{(\rho_s,a)\}$
        \ENDIF
    \ENDIF
\ENDFOR

\IF{$\mathcal{C}=\emptyset$}
    \RETURN $\bot$
\ENDIF

\STATE $(\rho^{st},a^{st})\leftarrow \texttt{Top}(\mathcal{C})$
\STATE $(\rho^{nd},a^{nd})\leftarrow \texttt{SecondTop}(\mathcal{C})$ if it exists

\IF{$a^{nd}$ exists and $a^{nd}\ne a^{st}$ and $\rho^{st}-\rho^{nd}<\lambda$}
    \RETURN $\bot$
\ENDIF

\RETURN $a^{st}$
\end{algorithmic}
\end{algorithm}

Therefore, a leaf-reflex rule that corresponds to multiple valid actions is treated as state ambiguity and rejected.
When multiple rules hit, the system uses only the support-derived score as a weak ranking signal.
If candidate actions differ and the top candidate does not exceed the second candidate by at least 1, the system abstains and returns to ReCAP.
On a hit, the unique valid action is used as the next environment action.

\subsection{Execution Write-Back and Fallback}

After a leaf-reflex hit, only the source of the leaf action changes; ReCAP's recursive state transition remains unchanged.
When the algorithm returns a unique valid action, HaReCAP treats it as a legal grounding result for the current leaf task and immediately interacts with the environment.
After the environment returns a new observation, control returns to the original ReCAP parent-node refinement process: the parent node updates the remaining plan, judges subtask progress, and determines the next recursive direction based on the latest feedback.
Thus, leaf-reflex triggering does not introduce an independent completion judgment, action repair, or error-recovery strategy.
When the triggering algorithm returns $\bot$, the system directly calls the original ReCAP leaf LLM; if the fallback LLM output still cannot be aligned with current valid actions, ReCAP's original failure handling and backtracking mechanism is used.

\section{Experiments}

\subsection{Experimental Setup}



The main experiments use Qwen3.5-27B on Robotouille \cite{robotouille} to evaluate the end-to-end success rate, support-threshold sensitivity, and paired-success reasoning cost of \method.
For Robotouille, the leaf-reflex library is built offline from 121 successful ReCAP baseline trajectories and contains 329 canonical leaf-reflex rules; at runtime, we ablate the minimum execution support threshold over 1, 2, and 4.

Based on this main setting, we further examine whether the mechanism depends on a single task domain or model capability level.
The cross-environment experiments are conducted on ALFWorld \cite{alfworld}, where a leaf-reflex library is rebuilt from successful ALFWorld trajectories using the same extraction procedure.
The cross-model experiments use Qwen3.5-9B and Gemma4-26B to answer whether the leaf-reflex mechanism depends on a single model scale or model family.
All within-group comparisons use low-temperature decoding, a context token limit of 30000, and a maximum generation length of 2048.
Metrics include success rate, LLM calls and token consumption.

We report both all-task and both-success-task results: the former measures end-to-end success over the full task set, while the latter compares execution efficiency only on tasks solved by both ReCAP and \method. This is important because long-horizon failures are often dominated by high-level planning drift or recovery loops, whose token costs are highly variable; both-success tasks provide more comparable trajectories for evaluating the cost impact of leaf-reflex grounding.

\subsection{Robotouille Main Results}

Table~\ref{tab:robo-success} reports end-to-end success rates on Robotouille with Qwen3.5-27B.
CoT and ReAct perform substantially below ReCAP, indicating that this environment requires recursive context management.
\method remains in a similar success-rate range to ReCAP on both the synchronous and asynchronous splits, suggesting that replacing leaf-level grounding does not disrupt ReCAP's high-level recursive control.

\begin{table}[t]
\caption{End-to-end success rates on Robotouille.}
\label{tab:robo-success}
\centering
\small
\begin{tabular*}{\linewidth}{@{\extracolsep{\fill}}lcc@{}}
\toprule
Method & Synchronous & Asynchronous \\
\midrule
CoT & 6\% & 0\% \\
ReAct & 22\% & 6\% \\
\midrule
ReCAP & 75\% & 46\% \\
\method\ s=1 & 76\% & 50\% \\
\method\ s=2 & 76\% & 48\% \\
\method\ s=4 & 74\% & 50\% \\
\bottomrule
\end{tabular*}
\end{table}


Table~\ref{tab:robo-both-success} further compares the reasoning cost of ReCAP and \method on Robotouille synchronous/asynchronous both-success tasks under Qwen3.5-27B.
Under the condition that the same tasks are completed by both methods, all three support thresholds reduce token consumption relative to ReCAP.
Support=2 achieves the largest reduction, lowering tokens by 14.67\% and 17.93\% on the synchronous and asynchronous splits, respectively.
Because the support threshold affects both rule coverage and rule reliability, support=1 covers more cases but includes more low-frequency experience, whereas support=4 is more conservative but reduces the number of triggerable rules.
Further increasing the threshold weakens leaf-reflex triggering and makes the system approach the original ReCAP execution mode.
Therefore, we use support=2 as the default setting for subsequent cross-environment and cross-model experiments.


These results show that part of ReCAP's repeated cost in long-horizon recursive control comes from last-mile action grounding.
This cost appears not only in relatively deterministic synchronous tasks, but also in asynchronous tasks with waiting and interleaved dependencies.
It can be replaced by leaf-reflex grounding to reduce inference cost without taking over high-level recursive planning.

\begin{table}[t]
\caption{Reasoning cost on Robotouille both-success tasks.}
\label{tab:robo-both-success}
\centering
\small
\newcommand{\drop}[1]{{\scriptsize\textcolor{green!50!black}{($\downarrow$#1)}}}
\newcommand{\tokval}[1]{\makebox[5.6em][r]{#1}\makebox[4.2em][l]{}}
\newcommand{\callval}[1]{\makebox[3.0em][r]{#1}\makebox[4.2em][l]{}}
\newcommand{\tokdrop}[2]{\makebox[5.6em][r]{#1}\makebox[4.2em][l]{\,\drop{#2}}}
\newcommand{\calldrop}[2]{\makebox[3.0em][r]{#1}\makebox[4.2em][l]{\,\drop{#2}}}
\renewcommand{\arraystretch}{1.06}
\begin{tabular*}{\linewidth}{@{\extracolsep{\fill}}cccccc@{}}
\toprule
Split & Support & Method & Both Succ. & Avg. Tok. & Avg. Calls \\
\midrule
\multirow{6}{*}{Sync}
& \multirow{2}{*}{$s=1$} & ReCAP & \multirow{2}{*}{64} & \tokval{753,455} & \callval{31.48} \\
& & \method & & \tokdrop{703,732}{6.60\%} & \calldrop{29.19}{7.30\%} \\
\cmidrule(lr){2-6}
& \multirow{2}{*}{$s=2$} & ReCAP & \multirow{2}{*}{64} & \tokval{772,053} & \callval{32.22} \\
& & \method & & \tokdrop{658,777}{14.67\%} & \calldrop{27.66}{14.15\%} \\
\cmidrule(lr){2-6}
& \multirow{2}{*}{$s=4$} & ReCAP & \multirow{2}{*}{63} & \tokval{747,628} & \callval{31.29} \\
& & \method & & \tokdrop{674,415}{9.79\%} & \calldrop{28.25}{9.72\%} \\
\midrule
\multirow{6}{*}{Async}
& \multirow{2}{*}{$s=1$} & ReCAP & \multirow{2}{*}{29} & \tokval{1,456,601} & \callval{56.38} \\
& & \method & & \tokdrop{1,248,787}{14.27\%} & \calldrop{49.28}{12.60\%} \\
\cmidrule(lr){2-6}
& \multirow{2}{*}{$s=2$} & ReCAP & \multirow{2}{*}{27} & \tokval{1,357,880} & \callval{52.85} \\
& & \method & & \tokdrop{1,114,477}{17.93\%} & \calldrop{44.22}{16.34\%} \\
\cmidrule(lr){2-6}
& \multirow{2}{*}{$s=4$} & ReCAP & \multirow{2}{*}{29} & \tokval{1,468,649} & \callval{56.93} \\
& & \method & & \tokdrop{1,351,911}{7.95\%} & \calldrop{53.28}{6.41\%} \\
\bottomrule
\end{tabular*}
\end{table}

\subsection{Cross-Environment and Cross-Model Generalization}

Robotouille verifies the efficiency gain of \method within a kitchen-like embodied environment.
To separate the portability of the method from task-domain-specific effects, we rebuild a leaf-reflex library on ALFWorld and repeat the evaluation across models.
ALFWorld differs from Robotouille in task semantics, interactive objects, and action space, so it tests whether leaf-reflex grounding can still replace repeated leaf action grounding under a new environment interface.
Qwen3.5-9B and Gemma4-26B are used to test whether the mechanism depends on Qwen3.5-27B.

Table~\ref{tab:alfworld-cross-env-model} reports the corresponding end-to-end results.
On Qwen3.5-27B, \method maintains a success rate close to ReCAP and reduces all-task average tokens by 20.7\%.
On Qwen3.5-9B and Gemma4-26B, \method also maintains comparable or higher success counts while reducing average tokens and LLM calls.

\begin{table}[t]
\caption{Cross-environment and cross-model results on ALFWorld.}
\label{tab:alfworld-cross-env-model}
\centering
\small
\providecommand{\drop}[1]{{\scriptsize\textcolor{green!50!black}{($\downarrow$#1)}}}
\providecommand{\tokval}[1]{\makebox[5.6em][r]{#1}\makebox[4.2em][l]{}}
\providecommand{\callval}[1]{\makebox[3.0em][r]{#1}\makebox[4.2em][l]{}}
\providecommand{\tokdrop}[2]{\makebox[5.6em][r]{#1}\makebox[4.2em][l]{\,\drop{#2}}}
\providecommand{\calldrop}[2]{\makebox[3.0em][r]{#1}\makebox[4.2em][l]{\,\drop{#2}}}
\renewcommand{\arraystretch}{1.06}
\begin{tabular*}{\linewidth}{@{\extracolsep{\fill}}ccccc@{}}
\toprule
Model & Method & Acc. & Avg. Tok. & Avg. Calls \\
\midrule
\multirow{2}{*}{Qwen3.5-27B}
& ReCAP & 94.03\% & \tokval{308,624} & \callval{29.90} \\
& \method\ s=2 & 93.28\% & \tokdrop{244,697}{20.7\%} & \calldrop{23.82}{20.3\%} \\
\cmidrule(lr){1-5}
\multirow{2}{*}{Qwen3.5-9B}
& ReCAP & 70.15\% & \tokval{348,457} & \callval{31.17} \\
& \method\ s=2 & 71.64\% & \tokdrop{282,035}{19.1\%} & \calldrop{25.24}{19.0\%} \\
\cmidrule(lr){1-5}
\multirow{2}{*}{Gemma4-26B}
& ReCAP & 66.42\% & \tokval{501,186} & \callval{43.84} \\
& \method\ s=2 & 75.37\% & \tokdrop{443,885}{11.4\%} & \calldrop{36.82}{16.0\%} \\
\bottomrule
\end{tabular*}
\end{table}

Table~\ref{tab:alfworld-paired} compares both-success tasks across different models on ALFWorld.
For all three model settings, \method reduces average total tokens and LLM calls on tasks solved by both methods.
The token reductions are 20.08\% for Qwen3.5-27B, 14.21\% for Qwen3.5-9B, and 21.09\% for Gemma4-26B.
Together with the Robotouille paired-success results, this shows that the core gain is not a kitchen-specific action rule, but a general replacement of repeated leaf grounding in ReCAP-style frameworks; the end-to-end behavior of leaf-reflex grounding also does not depend on a single model scale.

\begin{table}[t]
\caption{Reasoning cost on ALFWorld both-success tasks across models.}
\label{tab:alfworld-paired}
\centering
\small
\newcommand{\paireddrop}[1]{{\scriptsize\textcolor{green!50!black}{($\downarrow$#1)}}}
\newcommand{\pairedtokval}[1]{\makebox[5.6em][r]{#1}\makebox[4.2em][l]{}}
\newcommand{\pairedcallval}[1]{\makebox[3.0em][r]{#1}\makebox[4.2em][l]{}}
\newcommand{\pairedtokdrop}[2]{\makebox[5.6em][r]{#1}\makebox[4.2em][l]{\,\paireddrop{#2}}}
\newcommand{\pairedcalldrop}[2]{\makebox[3.0em][r]{#1}\makebox[4.2em][l]{\,\paireddrop{#2}}}
\renewcommand{\arraystretch}{1.06}
\begin{tabular*}{\linewidth}{@{\extracolsep{\fill}}ccccc@{}}
\toprule
Model & Method & Both Succ. & Avg. Tok. & Avg. Calls \\
\midrule
\multirow{2}{*}{Qwen3.5-27B}
& ReCAP & \multirow{2}{*}{123} & \pairedtokval{314,589} & \pairedcallval{30.25} \\
& \method\ s=2 & & \pairedtokdrop{251,411}{20.08\%} & \pairedcalldrop{24.22}{19.94\%} \\
\cmidrule(lr){1-5}
\multirow{2}{*}{Qwen3.5-9B}
& ReCAP & \multirow{2}{*}{83} & \pairedtokval{259,001} & \pairedcallval{27.43} \\
& \method\ s=2 & & \pairedtokdrop{222,207}{14.21\%} & \pairedcalldrop{22.65}{17.43\%} \\
\cmidrule(lr){1-5}
\multirow{2}{*}{Gemma4-26B}
& ReCAP & \multirow{2}{*}{82} & \pairedtokval{282,698} & \pairedcallval{30.15} \\
& \method\ s=2 & & \pairedtokdrop{223,069}{21.09\%} & \pairedcalldrop{24.09}{20.11\%} \\
\bottomrule
\end{tabular*}
\end{table}

Tables~\ref{tab:gemma4-robo-overall} and \ref{tab:gemma4-robo-paired} provide additional cross-model evidence on Robotouille.
Table~\ref{tab:gemma4-robo-overall} reports end-to-end success rates for Qwen3.5-9B and Gemma4-26B: although 9B performs substantially below 27B overall, \method does not cause systematic collapse; Gemma4-26B also remains close to ReCAP on both splits.
Table~\ref{tab:gemma4-robo-paired} compares Gemma4-26B on Robotouille both-success tasks and shows clear reductions in LLM calls and token consumption.
The results show that the HaReCAP tends to consume fewer tokens, and this phenomenon has certain generalizability, not being limited to a specific model or dataset.

\begin{table}[t]
\caption{Cross-model end-to-end success rates on Robotouille.}
\label{tab:gemma4-robo-overall}
\centering
\small
\begin{tabular*}{\linewidth}{@{\extracolsep{\fill}}llcc@{}}
\toprule
Model & Method & Synchronous & Asynchronous \\
\midrule
Qwen3.5-9B & ReCAP & 32\% & 5\% \\
Qwen3.5-9B & \method\ s=2 & 44\% & 10\% \\
\midrule
Gemma4-26B & ReCAP & 53\% & 26\% \\
Gemma4-26B & \method\ s=2 & 55\% & 24\% \\
\bottomrule
\end{tabular*}
\end{table}

\begin{table}[t]
\caption{Reasoning cost of Gemma4-26B on Robotouille both-success tasks.}
\label{tab:gemma4-robo-paired}
\centering
\small
\newcommand{\gemmadrop}[1]{{\scriptsize\textcolor{green!50!black}{($\downarrow$#1)}}}
\newcommand{\gemmatokval}[1]{\makebox[5.6em][r]{#1}\makebox[4.2em][l]{}}
\newcommand{\gemmacallval}[1]{\makebox[3.0em][r]{#1}\makebox[4.2em][l]{}}
\newcommand{\gemmatokdrop}[2]{\makebox[5.6em][r]{#1}\makebox[4.2em][l]{\,\gemmadrop{#2}}}
\newcommand{\gemmacalldrop}[2]{\makebox[3.0em][r]{#1}\makebox[4.2em][l]{\,\gemmadrop{#2}}}
\renewcommand{\arraystretch}{1.06}
\begin{tabular*}{\linewidth}{@{\extracolsep{\fill}}ccccc@{}}
\toprule
Split & Method & Both Succ. & Avg. Tok. & Avg. Calls \\
\midrule
\multirow{2}{*}{Sync}
& ReCAP & \multirow{2}{*}{44} & \gemmatokval{909,927} & \gemmacallval{38.66} \\
& \method\ s=2 & & \gemmatokdrop{611,890}{32.75\%} & \gemmacalldrop{26.14}{32.39\%} \\
\cmidrule(lr){1-5}
\multirow{2}{*}{Async}
& ReCAP & \multirow{2}{*}{10} & \gemmatokval{2,440,140} & \gemmacallval{92.60} \\
& \method\ s=2 & & \gemmatokdrop{2,000,589}{18.01\%} & \gemmacalldrop{76.20}{17.71\%} \\
\bottomrule
\end{tabular*}
\end{table}

\section{Conclusion}

This paper presents \method, a low-intrusion leaf-level action-grounding method for ReCAP-style recursive LLM agents.
It distills frequent leaf decisions from successful trajectories into leaf-reflex rules, and during execution triggers an action only when the rule uniquely matches the current valid-action set; otherwise it falls back to the original ReCAP.
Experiments on Robotouille and ALFWorld show that \method reduces token consumption and LLM calls on commonly successful tasks while maintaining similar end-to-end success rates.
The results indicate that part of the runtime cost in ReCAP-style recursive agents comes from repeated leaf action grounding rather than new high-level reasoning.
Turning these frequent, verifiable grounding decisions into abstainable leaf-reflex rules provides a lightweight, auditable, and low-intrusion path for improving the efficiency of long-horizon LLM agents.

\end{document}